\def\year{2020}\relax
\documentclass[letterpaper]{article} 
\usepackage{aaai20}  
\usepackage{times}  
\usepackage{helvet} 
\usepackage{courier}  
\usepackage[hyphens]{url}  
\usepackage{graphicx} 
\usepackage{graphicx}  
\title{Sanity Checks for Saliency Metrics}
\author{Richard Tomsett,\textsuperscript{\rm 1}\thanks{The first two authors contributed equally.} Dan Harborne,\textsuperscript{\rm 2}\footnotemark[1]
Supriyo Chakraborty,\textsuperscript{\rm 3}
Prudhvi Gurram,\textsuperscript{\rm 4}
Alun Preece\textsuperscript{\rm 2}\\ 
\textsuperscript{\rm 1}Emerging Technology, IBM Research, Hursley, UK \\
\textsuperscript{\rm 2}Crime and Security Research Institute, Cardiff University, Cardiff, UK \\
\textsuperscript{\rm 3}IBM Research, Yorktown Heights, NY, USA \\
\textsuperscript{\rm 4}Booz Allen Hamilton and CCDC Army Research Laboratory, Adelphi, MD, USA \\
{rtomsett@uk.ibm.com}, {harborned@cardiff.ac.uk},  {supriyo@us.ibm.com}, {gurram\_prudhvi@bah.com}, {preecead@cardiff.ac.uk}
}
 
\begin{document}

\maketitle

\begin{abstract}
Saliency maps are a popular approach to creating post-hoc explanations of image classifier outputs. These methods produce estimates of the relevance of each pixel to the classification output score, which can be displayed as a saliency map that highlights important pixels. Despite a proliferation of such methods, little effort has been made to quantify how good these saliency maps are at capturing the true relevance of the pixels to the classifier output (i.e. their ``fidelity''). We therefore investigate existing metrics for evaluating the fidelity of saliency methods (i.e. saliency metrics). We find that there is little consistency in the literature in how such metrics are calculated, and show that such inconsistencies can have a significant effect on the measured fidelity. Further, we apply measures of reliability developed in the psychometric testing literature to assess the consistency of saliency metrics when applied to individual saliency maps. Our results show that saliency metrics can be statistically unreliable and inconsistent, indicating that comparative rankings between saliency methods generated using such metrics can be untrustworthy.
\end{abstract}

\section{Introduction}
\label{sec:introduction}
Despite their popularity, deep neural networks (DNNs) are widely acknowledged to lack interpretability: their complex internals, with thousands or millions of parameters and non-linear elements, make it difficult to explain how and why a DNN maps inputs to outputs. This has led to many groups of researchers developing a variety of methods to help improve DNN interpretability. Saliency maps represent a popular class of explanation methods aimed at achieving this goal for DNNs operating on image data. They are designed to provide a measure of the relevance of each pixel to the DNN's output \cite{montavon2018review}. Many methods have been proposed in the recent past for generating saliency map type explanations~\cite{simonyan_deep_2013,zeiler_visualizing_2014,Bach2015lrp,montavon2017deeptaylor,ribeiro2016lime,Lundberg2018shap}. However, most of this prior work focuses on developing new methods to generate the saliency maps, without much work evaluating the quality of these methods. Different saliency methods provide different maps for the same image and corresponding DNN output, so human analysts need some way to decide which method provides the best estimates of the true pixel relevance values in order to choose between them.

Several authors have proposed axioms that explanation methods should adhere to in order to be produce appropriate explanations \cite{Sundararajan2017axiomatic,Kindermans2017}. These can be assessed by inspecting the mathematics behind the explanation methods, and working out if the method is consistent with the axioms. However, it is quite possible for useless explanation methods to be consistent with currently proposed axioms. For example, assigning uniform relevance scores to every pixel produces an explanation that is conservative, continuous, and implementation invariant~\cite{Montavon2019bookchapter}. To ensure that an explanation method actually assigns suitable relevance scores, we must assess whether the method truly discriminates appropriately between more and less relevant features in the input. This property has been referred to as the selectivity~\cite{Bach2015lrp} or fidelity~\cite{AlvarezMelis2018} of an explanation: how well it agrees with the way the model actually works. Put another way, a method with high fidelity will assign high relevance to features that, when removed, greatly reduce the DNN's output confidence in the class assignment, while assigning low relevance to features that do not greatly affect the confidence when removed. Fidelity cannot be established axiomatically, and so must be estimated using implementations of the explanation method with the DNN model and dataset under investigation.

In this paper, we investigate the properties of different approaches for measuring explanation method fidelity, focusing on saliency map explanations for image classifiers. We make the distinction between estimating the \textit{fidelity of an explanation method}, and estimating the \textit{fidelity of an individual explanation}. While currently proposed metrics assess the former, it is the latter that we are often interested in --- especially if we intend to use individual model outputs for decision making. Inspired by~\cite{Adebayo2018sanity}, who introduced a set of ``sanity checks'' (NB: \textit{not} metrics) for testing saliency methods, we propose a corresponding set of sanity checks for saliency \textit{metrics} based on measures of reliability from the psychometric testing literature. We use these checks to show that current metrics used to estimate saliency method fidelity can have high variance and are sensitive to implementation details. We also show that these metrics can be statistically unreliable when considered over individual saliency maps, and that the metrics we examined were not measuring the same underlying notion of fidelity.

\section{Background}
\label{sec: bg}
\subsection{Saliency methods}
\label{subsection:saliency}
Many methods have been proposed for improving the interpretability of neural networks (NNs). These can broadly be categorized into two types: (1) methods that investigate the NN's internals, and (2) methods that attempt to explain the NN output for individual inputs \cite{lipton2016mythos}. The former type of approaches are useful to researchers and engineers trying to understand their models, but less useful to a broader audience --- those without access to the model structure (e.g. those accessing a model-as-a-service via an API), or users without the relevant knowledge for interpreting information about NN internals. The latter approach of offering easily interpretable, ``post-hoc'' explanations \cite{lipton2016mythos} of outputs for specific inputs has therefore received much attention.

Feature attribution methods --- often called saliency maps when the features are pixels in images~\cite{Adebayo2018sanity} --- are a popular explanation technique for explaining classification outputs~\cite{montavon2018review}. These methods provide a measure of the relevance that each feature/pixel in a given input has to the model's output. Obtaining a correct value for the relevance of each feature is non-trivial: NNs are able to learn non-linear combinations of features that make separating their relative relevance to any particular output difficult. Additionally, many kinds of data have considerable local structure, and NN models are usually developed to take advantage of this. For example, pixels are only meaningful within an image when considered with nearby pixels, a property exploited by convolutional neural networks (CNNs).

Previously proposed methods for creating saliency maps for images have employed a variety of techniques for inferring pixel relevance~\cite{montavon2018review}. These have a variety of mathematical formulations that result in different properties for the saliency maps they produce. However, many are based on common theoretical grounds: for example, gradient-based methods rely on back-propagating gradients from output to input units to estimate how changes in the input will affect the output~\cite{selvaraju:gradcam}, while relevance-based methods estimate pixel relevance with reference to a root point in the input space~\cite{montavon2017deeptaylor}. Several authors have explored the theoretical relationships between different methods, finding many commonalities between different approaches~\cite{Lundberg2018shap,Ancona2017deepExplain,ancona2018ICLR}.

\subsection{Saliency metrics}
\label{subsection:metrics}
Explanation evaluation is inherently difficult because many different aspects of an explanation affect its perceived quality. One aspect of an explanation that should be quantifiable is the explanation's \textit{fidelity}: how well it genuinely represents the processing performed by a model on the input to produce the corresponding output. A measure of fidelity for saliency map methods should capture how well the method assigns relevance values to the input pixels.

Despite the large number of saliency methods available, relatively few metrics have been proposed for assessing their fidelity. One of the first approaches, initially outlined by \cite{Bach2015lrp} for binary images and extended by \cite{samek2017heatmapEvaluation} for RGB images, measures the change in classifier output as pixels are sequentially perturbed (flipped in binary images, or set to a different value for RGB images) in order of their relevance as estimated by the saliency method. The classification output should decrease more rapidly for methods that provide more accurate estimates of pixel relevance. Similarly, if the \textit{least} relevant pixels are perturbed first \cite{Arras2017rnn}, then the classification output should change more slowly the more accurate the saliency method. The proposed metric for capturing this is the Area Over the Perturbation Curve ($\mathrm{AOPC}_M$), which can be measured by perturbing pixels based either on the Most Relevant First (MoRF) or the Least Relevant First (LeRF) procedures:
\[
	\mathrm{AOPC}_M = \frac{1}{L+1} \left\langle \sum_{k=1}^L f(\mathbf{x}_{M}^{(0)}) - f(\mathbf{x}_{M}^{(k)}) \right\rangle_{p(\mathbf{x})},
\]

\noindent where $M$ is the pixel deletion procedure (MoRF or LeRF), $L$ is the number of pixel deletion steps, $f(\mathbf{x})$ is the output value of the classifier for input image $\mathbf{x}$ (i.e. the probability assigned to the highest-probability class), $\mathbf{x}_{M}^{(0)}$ is the input image after 0 perturbation steps (i.e. $\mathbf{x}_{M}^{(0)} = \mathbf{x}$), $\mathbf{x}_{M}^{(k)}$ is the input image after $k$ perturbation steps, and $\left\langle \cdot \right\rangle_{p(\mathbf{x})}$ denotes the mean over all images in the data set \cite{samek2017heatmapEvaluation}. Varying $L$ allows the measurement of $\mathrm{AOPC}_M$ over different amounts of pixel deletion.

A second metric described by \cite{AlvarezMelis2018} and simply called the ``faithfulness'', $\mathrm{F}$, similarly relies on removing features (perturbing pixels for images) and measuring the change in classification output. The metric as proposed in \cite{AlvarezMelis2018} differs from AOPC in that it perturbs pixels one-by-one, so that the change in output is only ever caused by a single pixel perturbation. The faithfulness $\mathrm{F}_{\mathbf{x}}$ for a single image $\mathbf{x}$ is calculated by taking the Pearson correlation between the relevance $R_i$ assigned to pixel $i$ by the saliency method, and the change in classification output when pixel $i$ is perturbed to create image $\mathbf{x}_i$: $\Delta_{i} = f(\mathbf{x}) - f(\mathbf{x}_{i})$, for all pixels in the image. The saliency method faithfulness, $\mathrm{F}$, is again calculated by taking the average faithfulness over all images in the data set:
\[
    \mathrm{F} = \left\langle \rho(R, \Delta) \right\rangle_{p(\mathbf{x})}.
\]

A third metric, RemOve And Retrain (ROAR) was proposed by \cite{ROAR2018}. The ROAR procedure calculates saliency maps for each image in the training data, perturbs the most relevant pixels, then retrains a new model with the same structure and initialization as the original model using the perturbed training data. If the saliency method identified the correct important pixels, the new classifier should exhibit a large reduction in accuracy compared to the original. Unlike the other proposed methods, ROAR does not allow for the measurement of individual saliency map fidelity --- only global saliency method fidelity. It thus cannot be used to assess the fidelity of individual saliency maps. Additionally, it requires extensive computational resources to retrain models from scratch after successive perturbation steps. We therefore did not consider ROAR further in our investigations.

\section{Evaluating saliency metrics}
\label{section:evaluating_metrics}

A metric should have good \textit{statistical validity}: it should measure the property that it is intended to measure. A true test of saliency metric validity would require knowledge of the ground-truth saliency maps --- which we do not have access to, and are what the saliency methods are trying to estimate.\footnote{In early experiments, we attempted to generate ground truth saliency maps by employing rules to generate images usually used in tests of human visual reasoning. However, we were still constrained to using very simple models to obtain these ground truth maps. Ultimately the results of these experiments were of limited use as they were not indicative of the results we obtained using the saliency methods and metrics with more complex models and datasets.} However, we can investigate a saliency metric's \textit{reliability} i.e. how well it provides consistent results. While a reliable metric is not necessarily valid, a valid metric \textit{must} be reliable \cite{davidshofer2005psychological}. 

To determine reliability, we consider metrics as applied to individual saliency maps, and make use of statistics usually used in the psychometric testing literature. We can think of a set of saliency methods as a battery of psychometric tests administered to an agent (the NN). The tests are scored by a saliency metric. Each input image corresponds to a different rater, who administers the battery of tests. Psychometric test reliability is usually estimated in four separate ways \cite{peter1979reliability}:

\begin{itemize}
  \item \textit{Inter-rater reliability}: degree to which different raters agree in their assessments. We use inter-rater reliability to refer to the agreement in the saliency metric scores between input images (raters) across different saliency methods, using the same metric.
  \item \textit{Inter-method reliability}: degree to which different tests agree in their assessments. We use inter-method reliability to refer to the agreement between different saliency methods (tests) across different input images, using the same metric.
  \item \textit{Internal consistency reliability}: degree to which different methods that are intended to measure the same concept produce similar scores. We use internal consistency reliability to refer to the agreement between different saliency metrics measuring the same saliency method.
  \item \textit{Test-retest reliability}: degree to which scores change between test administrations. Test-retest reliability is not relevant for saliency metrics because we are applying deterministic saliency methods to deterministic, fixed models.
\end{itemize}

Though this analogy is imperfect, it assists in the selection of appropriate statistics to test saliency metric reliability (our sanity checks), as described in the following subsections.

\subsection{Inter-rater reliability}
\label{subsection:inter-rater}
Using the above analogy, this class of reliability assesses how consistent a saliency metric is in its scores between different images. An initial assessment can be made by measuring the variance of the metric for a single saliency method over all data set images. However, a high variance may be acceptable as the variety of images in the test set may present a broad variety of difficulties for the saliency method. Instead, we can consider how consistent the metric is at ranking the fidelity of different saliency maps, when considered image by image. In other words, does the metric consistently rank some saliency methods higher than others over all the images?

The statistic commonly tested to answer this question is Krippendorf's $\alpha$, which is defined as
\[
    \alpha = 1 - \frac{D_o}{D_e},
\]
where $D_o$ is the observed disagreement in saliency method ranking between images, and $D_e$ is the disagreement expected by chance, taking into account the number of images and number of methods being ranked~\cite{krippendorff04}. Full details of how to calculate $D_o$ and $D_e$ are given in~\cite{krippendorff04}. If $\alpha = 1$, the saliency method ranking between images is totally consistent (i.e. the metric produces the same ranking over methods for every image), while $\alpha = 0$ implies the ranking between images is random (negative values indicate systematic disagreement). If the saliency metrics have a low value of $\alpha$, it implies that their ranking of saliency map fidelity for future test images is largely unpredictable.

\subsection{Inter-method reliability}
Inter-method reliability assesses whether a saliency metric agrees across different saliency methods. This can be measured by taking the pairwise correlations between the scores of the different saliency methods on the data set images. If the scores of each saliency method fluctuate similarly between images, these correlations will be high, indicating high inter-method reliability. We use Spearman's $\rho$ to measure these pair-wise correlations.

\subsection{Internal consistency reliability}
Internal consistency reliability indicates whether different saliency metrics are capturing the same underlying concept (saliency map fidelity). This can be measured by taking the correlation between the scores produced by different metrics over saliency maps produced by the same saliency method. Again, we use Spearman's $\rho$ to measure this.

\section{Experiments}
\label{section:experiments}
Our experiments were designed to measure the reliability of saliency metrics for saliency maps, as outlined in the previous section. We investigated both $\mathrm{AOPC}_M$ and faithfulness $\mathrm{F}$.

\subsection{Methods}
Our goal in this paper is to better understand saliency metrics, so instead of testing many saliency methods on many different models and tasks, we focused on a single model and a few saliency methods. This allowed us to explore the properties of the saliency metrics in depth, rather than being distracted by effects of different models and datasets on the metrics.

\subsubsection{Model and task}
We performed all our experiments on a CNN model trained on the CIFAR-10 dataset and classification task \cite{Krizhevsky2009CIFAR}. We chose CIFAR-10 because it is a well-known image classification set of suitable complexity (10 non-linearly separable classes) whose size (50,000 32$\times$32 RGB training images, 10,000 test images) is not prohibitive to running many experiments in a reasonable amount of time. We trained a standard CNN containing three sequential blocks, with each block consisting of two 2D convolution layers with batch normalization, followed by a max pooling layer. The output of the third block is flattened before connecting to the final classification layer. ReLUs were used for all activations except the classification layer, which used the standard SoftMax activation. The model had all bias terms set to zero, as the inclusion of bias terms can pose difficulties for relevance backpropagation approaches to pixel saliency estimation \cite{wang19bias,montavon2018review}. The model was trained on 45,000 training samples, with 5,000 held out as a validation set. During training the model was regularized using $l_2$ weight decay and dropout, and early stopping was used to prevent over-fitting. The resulting model achieved a test set accuracy of $86\%$.

\subsubsection{Saliency methods}
\begin{figure}[t]
\centering
\includegraphics[width=0.99\columnwidth]{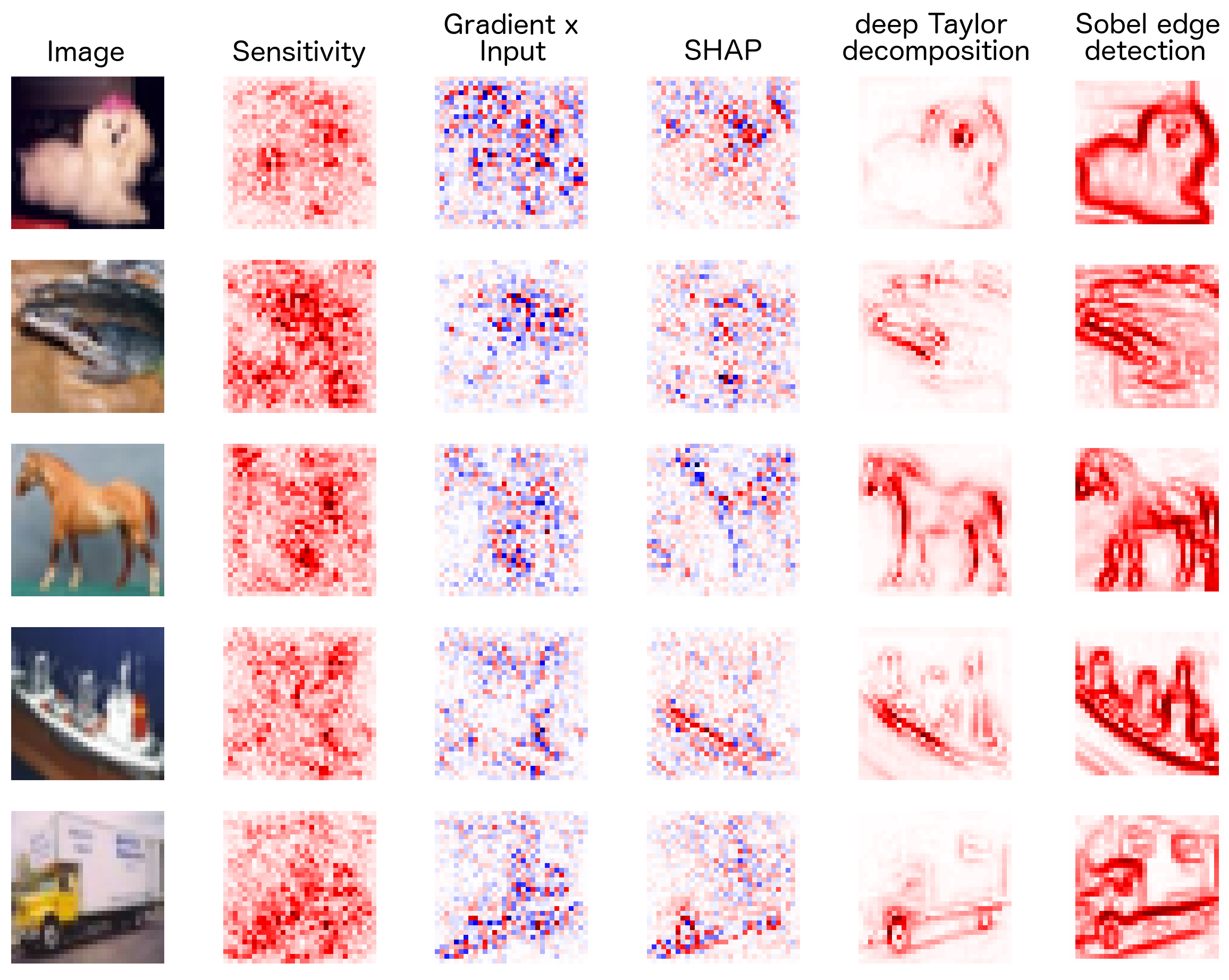} 
\caption{Example saliency maps using the four saliency methods plus edge detection, shown for five of the ten CIFAR-10 image classes (top to bottom: dog, frog, horse, ship, truck). Positive relevance shown in red, negative in blue. These images were all correctly classified by our NN with confidence $\geq 0.99$. Saliency maps were normalized by their maximum absolute value for visualization.}
\label{fig1}
\end{figure}

We chose to compare four different saliency methods and a baseline method (edge detection). Our aim here was not to be exhaustive, but to choose well-known methods from the literature with differing properties and assumptions. The methods chosen were sensitivity analysis~\cite{simonyan_deep_2013}, gradient$\odot$input,\footnote{It was noted by \cite{Kindermans2016gradientTimesInput,Shrikumar2017deeplift} that gradient$\odot$input was functionally equivalent to a version of the Layerwise Relevance Propagation method, $\epsilon$-LRP, \cite{Bach2015lrp,Kindermans2016gradientTimesInput} under certain conditions. We use gradient$\odot$input here as it is computationally simpler, and equivalent to $\epsilon$-LRP for our network.} deep Taylor decomposition~\cite{montavon2017deeptaylor}, and SHAP~\cite{Lundberg2018shap} --- specifically DeepSHAP, which builds on a connection between SHAP and DeepLIFT~\cite{Shrikumar2017deeplift}. Each of these produces a saliency map that assigns a relevance value to each pixel in the image, relative to the model's output classification score. Example saliency maps are shown in Figure~\ref{fig1}.

Some methods only estimate positive relevance i.e. all pixels are considered to have made a positive (or zero) contribution to the output classification, while other methods estimate positive \textit{and} negative relevance. Of the methods we looked at, sensitivity analysis and deep Taylor decomposition estimate positive relevance only, while gradient$\odot$input and SHAP estimate positive and negative relevance. We use the implementations of gradient (which we convert to sensitivity by taking the channel-wise maximum of the magnitude of the gradient), gradient$\odot$input, and deep Taylor decomposition provided by the iNNvestigate toolbox \cite{Alber2019iNNvestigate}, and the implementation of DeepSHAP available at https://github.com/slundberg/shap/ . Finally, we also compare these methods with saliency maps produced by Sobel edge detection, which acts as a baseline. Edge detection creates maps that are visually similar to those produced several saliency methods, but that do not depend on the NN internals~\cite{Adebayo2018sanity}.

\subsubsection{Saliency metrics}
\label{subsubsection:saliency_metrics}
We investigated AOPC with both MoRF and LeRF pixel perturbation order, and faithfulness $\mathrm{F}$. These metrics require pixels to be ``perturbed'' but for colour images, the correct perturbation function is not obvious. Several different approaches have been proposed in the literature, largely without explicitly stating why that particular function was chosen. We use two different perturbations: replacing the selected pixel with the dataset mean, and replacing the selected pixel with uniformly distributed random RGB values. The former approach effectively sets the pixels to 0 as we standardize inputs to the network, while the latter attempts to destroy the information contained in the pixel, as well as its correlation with surrounding pixels. Some previous studies have chosen to perturb square pixel-regions rather than individual pixels~\cite{samek2017heatmapEvaluation}; we chose to perturb single pixels as perturbing pixel regions requires assumptions about the spatial scale of the features in the images and the information contained in the saliency maps. We also note that perturbing a single pixel in a CIFAR10 image is similar to perturbing a $9\times9$ region in an ImageNet~\cite{imagenet_cvpr09} image as done by e.g.~\cite{samek2017heatmapEvaluation} (1 pixel is $~0.1\%$ of a CIFAR10 image's area; a $9\times9$ region is $~0.16\%$ of an ImageNet image's area assuming standard resizing to $224\times224$).

For the faithfulness results, we did not perturb every pixel in every image as this would have been computationally prohibitive. Instead we randomly selected 100 pixel IDs and perturbed this same pixel set for every image to obtain an estimate of the faithfulness. For the AOPC results, we also scored a set of random perturbation orderings to obtain a random baseline in addition to the edge detection baseline. We generated 100 random pixel orderings, using this set to obtain a set of 100 AOPC scores for every image. We use the mean of these 100 scores as the random baseline score, with a 95\% confidence interval calculated on the empirical distribution of mean scores. Other confidence intervals were estimated using the bootstrap method with 10,000 re-samplings. We performed our experiments on the whole of the CIFAR-10 test set of 10,000 images (1000 per class).

\section{Results}
\label{section:results}

\subsection{Global saliency metric reliability}
\begin{figure*}[t!]
\centering
\includegraphics[width=0.99\textwidth]{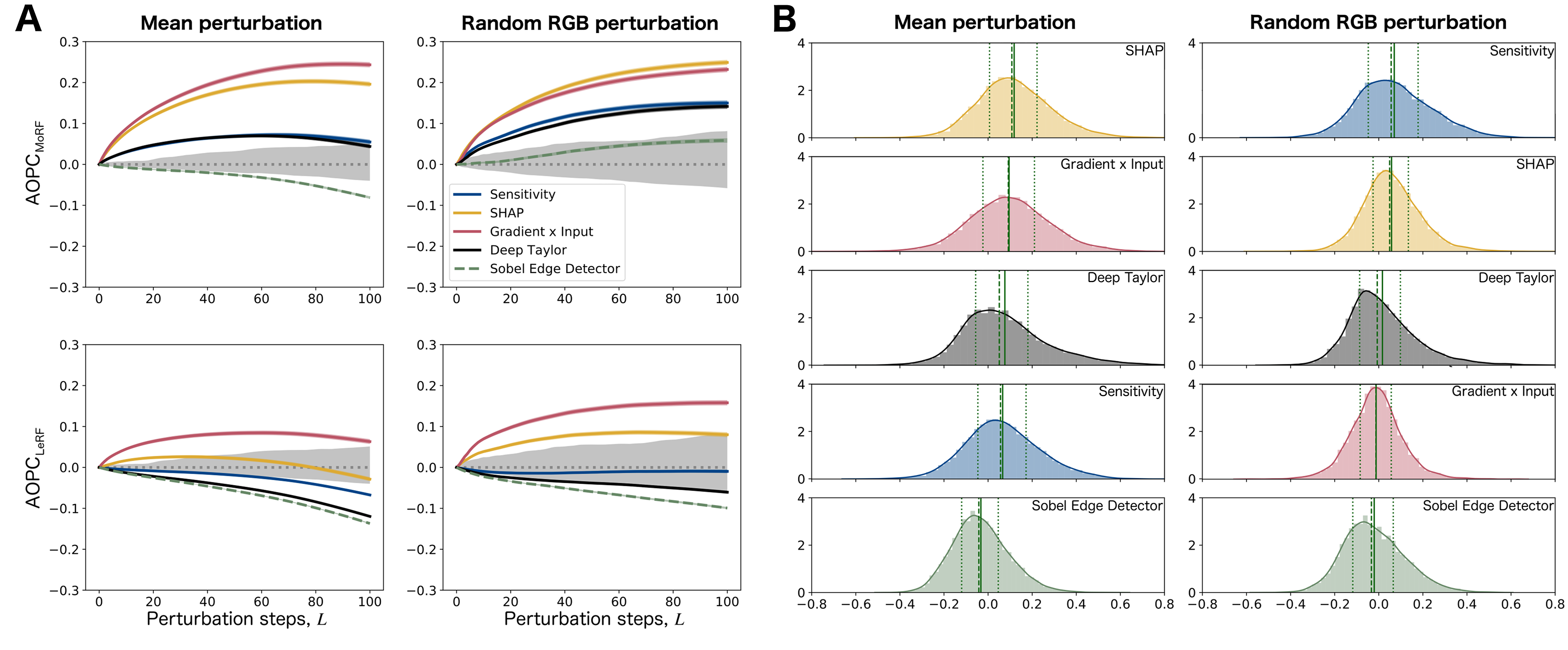} 
\caption{Metrics for saliency methods. \textbf{(a)} AOPC scores for each saliency method, measured from $L=0$ to $L=100$ perturbation steps. Scores adjusted by subtracting the mean AOPC for random deletion (dotted line with grey shading). \textit{Left column}: results using mean perturbation. \textit{Right colum}: results using random RGB perturbation. \textit{Top row:} Results for MoRF pixel order. Higher $\mathrm{AOPC_{MoRF}}$ is better. \textit{Bottom row}: results for LeRF pixel order. Lower $\mathrm{AOPC_{LeRF}}$ is better. 95\% confidence intervals shaded (very tight, so barely visible except for the random baseline). \textbf{(b)} Faithfulness ($\mathrm{F}$) scores for each saliency method. Distribution over all images is show as a histogram and density plot. Mean shown as vertical solid line; median is vertical dashed line; upper and lower quartiles are vertical dotted lines. Rows are ordered by metric rank (highest mean $\mathrm{F}$ top row, to lowest mean $\mathrm{F}$ bottom row. \textit{Left column}: results using mean perturbation. \textit{Right column}: results using random RGB perturbation.}
\label{fig2}
\end{figure*}

Figure \ref{fig2} shows the two metrics as measured for each saliency method, obtained using both mean perturbation (setting perturbed pixels to the data set mean value) and random RGB perturbation (setting perturbed pixels to uniformly random RGB values). Figure \ref{fig2}a shows AOPC values plotted against number of deletion steps; the top row shows results for MoRF perturbation and the bottom row for LeRF. These plots reveal several details about the reliability of AOPC measures. Considering first MoRF perturbation order (where higher AOPC values indicate better fidelity), gradient$\odot$input and SHAP are the top-ranked methods for mean and random RGB perturbation, respectively. Deep Taylor decomposition and sensitivity are indistinguishable using mean perturbation, but sensitivity is ranked better than deep Taylor decomposition with random RGB perturbation. This inconsistency indicates that $\mathrm{AOPC_{MoRF}}$ is sensitive to the details of the perturbation function for our CNN. Turning to LeRF perturbation order (where lower AOPC values indicate better fidelity), the saliency methods are ranked consistently between mean and random RGB perturbation. However, the LeRF rankings are opposite to those measured by MoRF: edge detection and deep Taylor decomposition obtain the best scores, with SHAP third and gradient$\odot$input last. This indicates that $\mathrm{AOPC_{MoRF}}$ and $\mathrm{AOPC_{LeRF}}$ are not measuring the same thing (low internal consistency reliability). A final thing to note regarding the plots in Figure \ref{fig2}a is the hidden variance of the AOPC measures: the 95\% confidence intervals for the means are very tight due to the large sample size, but the variance in AOPC scores over input images is large.

Figure \ref{fig2}b shows Faithfulness, $\mathrm{F}$, for each saliency method. In these plots we show the full distribution of $\mathrm{F}$ scores measured on each image, as well as the means indicating the global scores for the different methods. The mean values for every method are close to zero, indicating that they all have low faithfulness, with edge detection consistently the lowest. Again, the perturbation method affects the metric, changing, the distribution shapes, mean values, and method rankings.

\subsection{Local saliency metric reliability}
\begin{table*}[ht!]
\centering
\begin{tabular}{rcccc}
\hline \hline
\multicolumn{1}{l}{} & \begin{tabular}[c]{@{}c@{}}Mean perturb,\\ with edge detection\end{tabular} & \begin{tabular}[c]{@{}c@{}}Random perturb,\\ with edge detection\end{tabular} & \begin{tabular}[c]{@{}c@{}}Mean perturb,\\ excluding edge detection\end{tabular} & \begin{tabular}[c]{@{}c@{}}Random perturb,\\ excluding edge detection\end{tabular} \\ \hline \hline
Faithfulness, $\mathrm{F}$ & 0.18 (0.17---0.20) & 0.10 (0.09---0.12) & 0.04 (0.03---0.05) & 0.09 (0.08---0.10) \\ \hline
\begin{tabular}[c]{@{}r@{}}$\mathrm{AOPC_{MoRF}}$,\\ $L$=20\end{tabular} & 0.29 (0.27---0.31) & 0.15 (0.14---0.17) & 0.19 (0.17---0.20) & 0.07 (0.06---0.08) \\ \hline
\begin{tabular}[c]{@{}r@{}}$\mathrm{AOPC_{MoRF}}$,\\ $L$=40\end{tabular} & 0.38 (0.36---0.39) & 0.18 (0.17---0.20) & 0.27 (0.25---0.28) & 0.09 (0.08---0.11) \\ \hline
\begin{tabular}[c]{@{}r@{}}$\mathrm{AOPC_{MoRF}}$,\\ $L$=60\end{tabular} & 0.42 (0.40---0.44) & 0.20 (0.18---0.21) & 0.31 (0.29---0.33) & 0.11 (0.10---0.12) \\ \hline
\begin{tabular}[c]{@{}r@{}}$\mathrm{AOPC_{MoRF}}$,\\ $L$=80\end{tabular} & 0.46 (0.44---0.48) & 0.21 (0.20---0.23) & 0.34 (0.32---0.36) & 0.12 (0.11---0.14) \\ \hline
\begin{tabular}[c]{@{}r@{}}$\mathrm{AOPC_{MoRF}}$,\\ $L$=100\end{tabular} & 0.48 (0.46---0.50) & 0.22 (0.21---0.24) & 0.37 (0.35---0.39) & 0.13 (0.12---0.15) \\ \hline \hline
\begin{tabular}[c]{@{}r@{}}$\mathrm{AOPC_{LeRF}}$,\\ $L$=20\end{tabular} & 0.05 (0.04---0.06) & 0.13 (0.11---0.15) & 0.06 (0.05---0.07) & 0.15 (0.13---0.17) \\ \hline
\begin{tabular}[c]{@{}r@{}}$\mathrm{AOPC_{LeRF}}$,\\ $L$=40\end{tabular} & 0.11 (0.09---0.12) & 0.18 (0.16---0.20) & 0.11 (0.09---0.12) & 0.20 (0.18---0.22) \\ \hline
\begin{tabular}[c]{@{}r@{}}$\mathrm{AOPC_{LeRF}}$,\\ $L$=60\end{tabular} & 0.16 (0.14---0.17) & 0.22 (0.20---0.24) & 0.15 (0.14---0.17) & 0.23 (0.21---0.25) \\ \hline
\begin{tabular}[c]{@{}r@{}}$\mathrm{AOPC_{LeRF}}$,\\ $L$=80\end{tabular} & 0.20 (0.18---0.22) & 0.25 (0.23---0.27) & 0.19 (0.17---0.21) & 0.26 (0.23---0.28) \\ \hline
\begin{tabular}[c]{@{}r@{}}$\mathrm{AOPC_{LeRF}}$,\\ $L$=100\end{tabular} & 0.24 (0.22---0.25) & 0.28 (0.26---0.30) & 0.23 (0.21---0.25) & 0.28 (0.26---0.30) \\ \hline
\end{tabular}
\caption{Krippendorf's $\alpha$ under different perturbation functions for the saliency metrics, measured using saliency method ranking across images. Numbers in brackets are 99.9\% confidence intervals estimated with 10,000 bootstrap samples.}
\label{tab:krippendorf}
\end{table*}

\begin{table*}[ht!]
\centering
\begin{tabular}{cccccc}
\hline
 \multicolumn{1}{c}{} & Sensitivity & gradient$\odot$input & SHAP & Deep Taylor & Edge detection \\ \hline
\begin{tabular}[c]{@{}r@{}}Faithfulness, $\mathrm{F}$\\ vs\\ $\mathrm{AOPC_{MoRF}}$, \end{tabular} & 0.34 & 0.11 & 0.00 & 0.24 & 0.17 \\ \hline
\begin{tabular}[c]{@{}r@{}}Faithfulness\\ vs\\ $\mathrm{AOPC_{LeRF}}$, \end{tabular} & -0.11 & -0.14 & -0.09 & -0.22 & -0.09 \\ \hline
\begin{tabular}[c]{@{}r@{}}$\mathrm{AOPC_{MoRF}}$\\ vs\\ $\mathrm{AOPC_{LeRF}}$ \end{tabular} & 0.28 & 0.53 & 0.77 & 0.16 & 0.11 \\ \hline
\begin{tabular}[c]{@{}r@{}}$\mathrm{AOPC_{MoRF}}$, random RGB perturb\\ vs\\ $\mathrm{AOPC_{MoRF}}$, mean perturb\end{tabular} & 0.62 & 0.71 & 0.77 & 0.58 & 0.57 \\ \hline
\end{tabular}
\caption{Spearman correlations between pairs of metrics, measured over all images, for each saliency method. AOPC was taken at $L=100$ perturbation steps. Unless otherwise stated, mean perturbation was used.}
\label{tab:consistency}
\end{table*}

Table \ref{tab:krippendorf} lists the Krippendorf $\alpha$ statistics for the saliency metrics (with varying numbers of deleted pixels $L$ for AOPC). These are calculated on the image-wise rankings of saliency maps, and test ``inter-rater reliability'' as described above. Low $\alpha$ values indicate that the saliency method rankings on different images are inconsistent. The left two columns of table \ref{tab:krippendorf} show $\alpha$ values when the baseline edge detection method is included in the ranking. This baseline is ranked more consistently (low rankings for $\mathrm{AOPC_{MoRF}}$ and $\mathrm{F}$, high rankings for $\mathrm{AOPC_{LeRF}}$) than the true saliency methods over all data set images, producing higher $\alpha$ values than when it is excluded from the rankings (the right-hand two columns).

Perturbing with random RGB values reduces $\alpha$ for Fidelity and $\mathrm{AOPC_{MoRF}}$ compared with mean perturbation, which may be due to the increased stochasticity in the perturbations from the random colour choices. However, it is difficult to understand why $\mathrm{AOPC_{LeRF}}$ at smaller numbers of perturbation steps ($L=20$ and $L=40$) produce greater $\alpha$ values with random RGB perturbation that with mean perturbation. A ``low'' $\alpha$ value is not strictly defined, but $\alpha<0.65$ are often considered to indicate unreliability inter-rater reliability. The largest value in the table, 0.48 for $\mathrm{AOPC_{MoRF}}$ at $L=100$ using mean perturbation and including edge detection in the ranking, is substantially less than 0.65, indicating low inter-rater reliability whatever the specifics of the metric. We can conclude from this that the metrics will not produce consistent rankings of saliency maps when applied to new test images.

We made a similar set of measurements looking at the inter-method correlation of metric scores over data set images, taking the mean of the pair-wise Spearman's $\rho$ correlation between saliency methods. As with the previous results, the presence of the baseline edge detector --- which should not correlate well with other methods given that it does not assess the model --- reduces the mean pairwise correlation. The lowest mean pairwise correlation was 0.13 (for Faithfulness, random RGB perturbation, edge detection excluded) and the highest 0.67 (for $\mathrm{AOPC_{MoRF}}$, $L=100$, mean perturbation, edge detection excluded). Correlations for $\mathrm{AOPC_{MoRF}}$ were consistently highest, ranging between 0.47 and 0.67, while Faithfulness showed the lowest correlation values of 0.13 to 0.18 (both excluding edge detection). Changes in perturbation method appear to affect inter-method reliability much less than inter-rater reliability.

Finally, we assess internal consistency reliability by measuring the correlation between different metrics over the data set images, for each saliency method. A selection of pairs of correlations for different metrics is shown in table \ref{tab:consistency}. The correlation values indicate that the metrics are, in general, not measuring the same underlying quantity across the different saliency methods. Faithfulness is only weakly correlated with $\mathrm{AOPC_{MoRF}}$, and in fact slightly anti-correlated with $\mathrm{AOPC_{LeRF}}$ across all saliency methods. The correlation between $\mathrm{AOPC_{MoRF}}$ and $\mathrm{AOPC_{LeRF}}$ is highly variable across saliency methods, while the most consistent correlations across methods are between $\mathrm{AOPC_{MoRF}}$ applied with random RGB versus mean perturbation. This is to be expected as both methods perturb the same pixels --- though there is still reasonably high variability even in this case (from $\rho=0.58$ for deep Taylor decomposition to $\rho=0.77$ for SHAP).

\section{Discussion}
\label{sec:discussion}

We investigated the reliability of several saliency metrics that have been proposed to measure saliency method fidelity. To do this, we applied the metrics to a single model and task, assessing the reliability of the metrics in relation to this model. It is important to note that our results and conclusions are not generally applicable statements about the consistency of saliency metrics --- they are an illustration that such metrics \textit{can} produce unreliable and inconsistent results. Users should be wary of these properties when applying saliency metrics, and we suggest they apply the above sanity checks to any metrics in the context of their particular data and model.

Our main findings can be summarized thus:

\begin{itemize}
  \item \textit{Global saliency metrics had high variance}: The distribution of scores over all test images for each saliency metric was broad. Past work has only reported mean metric values \cite{Arras2017rnn,samek2017heatmapEvaluation,Bach2015lrp,montavon2018review}, ignoring this variance.
  \item \textit{Saliency metrics were sensitive to the specifics of their implementation}: we observed that both the faithfulness and AOPC scores, and the ranking of the saliency methods according to these scores, were affected by the specifics of the perturbation method (in our case, mean vs random RGB perturbation).
  \item \textit{Saliency maps from different saliency methods were ranked inconsistently image-by-image}: each metric exhibited low inter-rater reliability, meaning that rankings of saliency method fidelity considered image-by-image were highly inconsistent. This suggests that global saliency metric scores provide an unreliable indication of the fidelity of different saliency methods on future test images.
  \item \textit{The internal consistency of different metrics that all attempt to measure fidelity was low}: the correlation between different metric scores over saliency maps is generally low, and highly variable. This means that the different metrics we tested were not necessarily measuring the same underlying concept.
\end{itemize}

The above findings point to the great difficulty in measuring saliency map fidelity when we do not have access to ground truth saliency maps. In particular, it is very hard to disentangle the sources of variance and inconsistency in the metric scores, as model, saliency method and metric are tightly coupled. The specifics of the pixel perturbation method are clearly important as demonstrated above: however carefully implemented, these have the potential to move the image off the data manifold learned by the NN, thus reducing the validity of any measurements. \cite{montavon2018review} recommend choosing a perturbation method that is less likely to move the image off-manifold, but it is not currently possible to know what method will adhere to this requirement a priori.

Another notable observation was the reversal in global saliency method fidelity rankings between $\mathrm{AOPC_{MoRF}}$ and $\mathrm{AOPC_{LeRF}}$. Two effects appeared to contribute to this. First, LeRF perturbation order apparently favors methods that are more edge-detection like: edge detection and deep Taylor decomposition perform best with LeRF. These methods assign low relevance to large, contiguous blocks of color, and setting these regions to the dataset mean removes little discriminative information from the image. Equally, mean-perturbing all pixels along an edge does not remove the edge, it merely changes its contrast. The information in high-relevance pixels is therefore not fully removed by mean perturbation, as illustrated by deep Taylor decomposition's poor performance under MoRF. Also note that edge detection and deep Taylor decomposition’s $\mathrm{AOPC_{MoRF}}$ scores with random RGB perturbation are better than with mean perturbation, as random perturbation removes edge information more successfully. Second, gradient$\odot$input and SHAP --- which produce both positive and negative relevance values --- perform worse than random under LeRF (when the most negative pixels are perturbed first, rather than the pixels with the lowest absolute values) but do well under MoRF. We noted that the model was very confident on most inputs (59\% of test images classified with confidence $\geq0.99$) despite the presence of negative relevance pixels in gradient$\odot$input and SHAP saliency maps. When we considered only the 1600 images classified with confidence $\leq0.8$, LeRF scores were better than random for the methods producing negative relevance. This suggests negative relevance in saliency maps for high-confidence classifications contribute to poor LeRF scores.

Further insights into the underlying causes of these results could be gleaned by considering the metric results on each image class individually. We found that all metric scores were dependent on image class; e.g. saliency maps of frog images produced generally better $\mathrm{AOPC_{MoRF}}$ scores, worse $\mathrm{AOPC_{LeRF}}$ scores, and better $\mathrm{F}$ scores than other classes. The reasons for this are yet to be fully examined, but will depend on the systematic differences in the structure of the higher-level features in different classes, how the classifier has learned to represent these features, and how well the saliency maps capture these representations. We leave a fuller analysis along these lines as future work.

\section{Conclusion}
\label{sec:conclusion}
Inspired by \cite{Adebayo2018sanity}'s sanity checks for saliency maps, we investigated a set of sanity checks for saliency \textit{metrics}. To be useful, metrics should have high statistical validity. As this is not possible to establish without (unavailable) ground truth references, we can at least investigate metric \textit{reliability}. We illustrated the application of different measures of reliability to saliency metrics, specifically inter-rater reliability, inter-method reliability, and internal consistency. These sanity checks showed that current metrics were unreliable at measuring saliency map fidelity for our model. Due to the tight coupling of model, saliency method and saliency metric, it will be very challenging to disentangle the causes of this unreliability.

Based on our results, we propose the following recommendations for future metric development:

\begin{itemize}
  \item New metrics should be compared directly with previous metrics purporting to measure the same or related quantities, and their differences made explicit (preferably quantitatively). 
  \item If the metric could be implemented in different ways (e.g. mean versus random RGB pixel perturbation), the properties of these metric variants should be analyzed under different conditions to establish the effects of these different implementations.
  \item Metrics should be analyzed for how they might be ``tricked'' in different contexts (e.g. edge detection giving the best $\mathrm{AOPC_{LeRF}}$ scores)
  \item Metric developers should encourage users of their metric to investigate and understand the sources of variance in the metric scores, and how this affects their decisions about what saliency methods to choose for their particular model.
\end{itemize}

We also suggest that developers of saliency methods should not rely on a single metric to test their method's fidelity, as the results could be misleading. Instead, they should employ several metrics and attempt to understand sources of unreliability for their method on a variety of datasets and models.

\section{Acknowledgments}
We thank the anonymous reviewers for their comments, which were very helpful in improving the final manuscript.

This research was sponsored by the U.S. Army Research Laboratory and the UK Ministry of Defence under Agreement Number W911NF-16-3-0001. The views and conclusions contained in this document are those of the authors and should not be interpreted as representing the official policies, either expressed or implied, of the U.S. Army Research Laboratory, the U.S. Government, the UK Ministry of Defence or the UK Government. The U.S. and UK Governments are authorized to reproduce and distribute reprints for Government purposes notwithstanding any copy-right notation hereon.

\bibliographystyle{aaai}
\bibliography{AAAI-TomsettR.8794}
\end{document}